\documentclass{ws-rv9x6}
\usepackage{subfigure}     % required only when side-by-side / subfigures are used
\usepackage{ws-rv-van}     % numbered citation/references (default)
\makeindex
\begin{document}

\chapter * {Classification and its applications for drug-target interaction identification}%\label{ra_ch1}
%\title{Classification and its application to drug-target interaction prediction}%\label{ra_ch1}

\author[Jian-Ping Mei, Chee-Keong Kwoh, Peng Yang and Xiao-Li Li]{Jian-Ping Mei, Chee-Keong Kwoh, \\ %\footnote{Author footnote.},
Peng Yang and Xiao-Li Li}
%\index[aindx]{Author, F.} % or \aindx{Author, F.}
%\index[aindx]{Author, S.} % or \aindx{Author, S.}

\address{School of Computer Science, Nanyang Technological University\\
50 Nanyang Avenue, Singapore 639798.
%\footnote{Affiliation footnote.}
}
%\index[aindx]{Author, F.} % or \aindx{Author, F.}
%\index[aindx]{Author, S.} % or \aindx{Author, S.}

\begin{abstract}
Classification is one of the most popular and widely used supervised learning tasks, which categorizes objects into predefined classes based on known knowledge. Classification has been an important research topic in machine learning and data mining. Different classification methods have been proposed and applied to deal with various real-world problems. Unlike unsupervised learning such as clustering, a classifier is typically trained with labeled data before being used to make prediction, and usually achieves higher accuracy than unsupervised one.

In this chapter, we first define classification and then review several representative methods. After that, we study in details the application of classification to a critical problem in drug discovery, i.e., drug-target prediction, due to the challenges in predicting possible interactions between drugs and targets.

\end{abstract}
%\markright{Customized Running Head for Odd Page} % default is chapter title.
\body

\section{Classification}\label{ra_sec1}
Classification is the process of finding a model or function that describes and distinguishes data classes or concepts \citep{hanjiawei}. It is one of the most important tasks that supervised learning is applied to.
Supervised learning is an important machine learning method which learns a model or a function with the help of supervision.  Other than classification, supervised learning is also used for regression analysis. The goal of classification analysis is simply to know the class label while regression analysis is to learn a function.

The rapid development of technologies, such as microarrays,
 high-throughput sequencing, genotyping
 arrays, mass spectrometry, and automated
 high-resolution imaging acquisition techniques, has
 led to a dramatic increase in availability of biomedical
 data \citep{chen11classiftree}. Facing large amount of data, computational method, which is cheaper and more efficient, arises to be useful complement to support traditional experimental method in many biomedical researches and applications. As an important data analysis tool, classification has been applied for handling many important tasks in bioinformatics, including Sequence annotation \citep{EDWARD91SeqNN,Salzberg95SeqDT,Gupta10}, Protein function prediction \citep{Borgwardt05profunc}, Protein structure prediction \citep{McLaughlin03}, Gene regulatory network inference \citep{Mordelet08,Cerulo10}, Protein-protein interaction prediction \citep{qi06ppi}, disease gene identification \citep{yang12,yang2011inferring, yang2014ensemble} and drug-target interaction prediction \citep{Yamanishi08,Bleakley09,Mei12,mei2013drug}.
Many of these tasks are to search the answer of a question with ``yes'' or ``no''. For example, to predict whether two proteins interact or not, a protein is enzyme or non enzyme, a piece of sequence is coding or non-coding, and so on. This type of prediction can be directly handled with binary classification, where ``yes'' and ``no'' are treated as two class labels. It also can be solved through regression methods. Instead of directly answer ``yes'' or ``no'', binomial regression methods produce the likelihood or the degree of being ``yes'' or ``no'', based on which the final result can be easily obtained by cutting with a certain threshold, i.e., ``yes'' if the likelihood is larger than the threshold, and ``no'' if the likelihood value is below the given threshold. Next we give more detailed introduction on several representative classification methods, which are most widely used in computational biology.

In the following subsections, we represent the training data consisting of $n$ labeled examples or data objects as $D=\{\textbf{x}_i, y_i\}^n_{i=1}$, where each $\textbf{x}_i$ is a $p$-dimensional vector, i.e., $\textbf{x}_i=(x_{i1} \ldots x_{ip})^T$ and $y_i$ is its associated class label.
\subsection{k-Nearest Neighbor (K-NN)}
k-Nearest Neighbor (K-NN) is instance-based classification. In K-NN, an unlabeled object is assigned to the most common class among its $k$ most nearest neighbors in the training set. In order to decide the $k$ nearest neighbors of the given object, the distance or closeness between this object and all the labeled objects need to be calculated. The number of neighbors $k$ is an important parameter in k-NN. Setting $k$ to different values, k-NN may produce different results.

\begin{figure}[!tpb]%figure1
\centering
\includegraphics[width=0.5\textwidth]{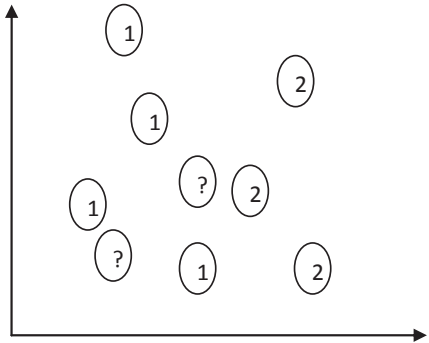}
\caption{A simple two-dimensional dataset.
}\label{fig:knndata}
\end{figure}

\begin{figure}[!tpb]%figure1
\centering
\includegraphics[width=0.9\textwidth]{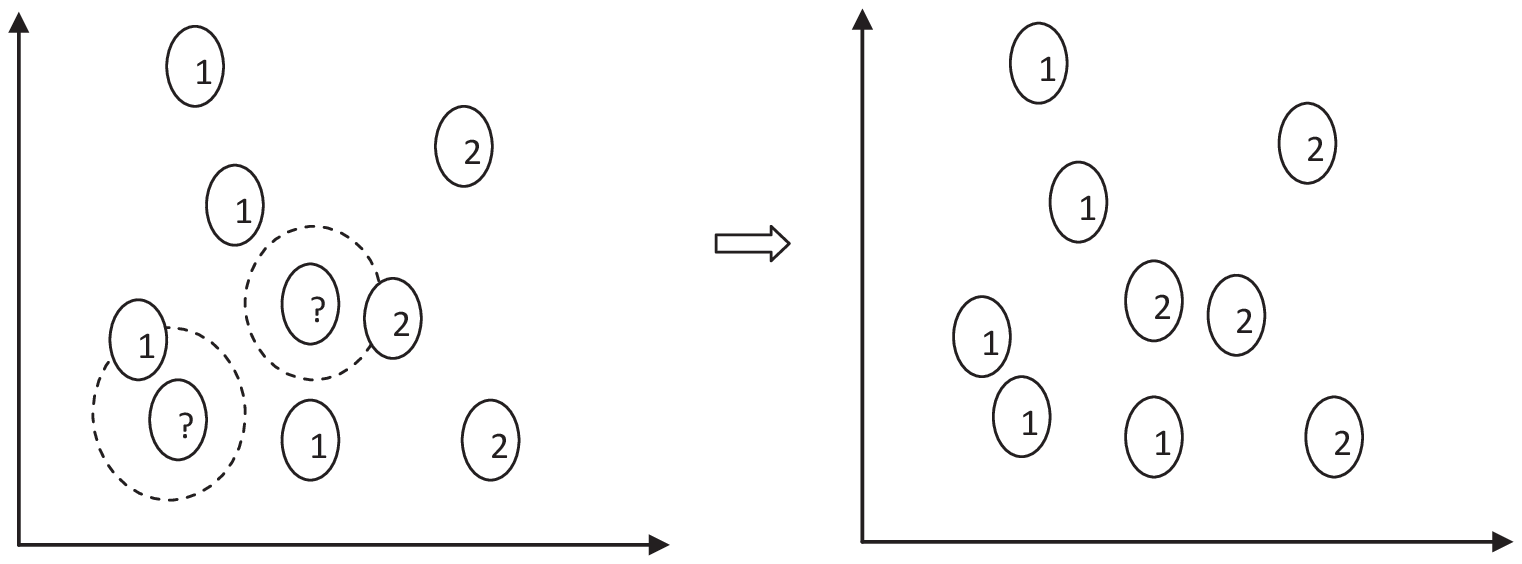}
\caption{Classification result of k-NN with $k=1$.
}\label{fig:knn_k=1}
\end{figure}

\begin{figure}[!tpb]%figure1
\centering
\includegraphics[width=0.9\textwidth]{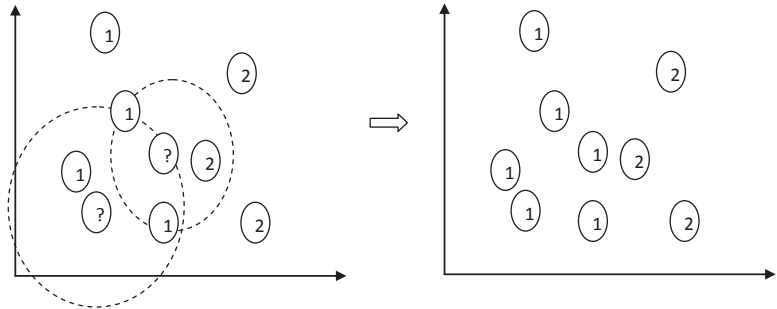}
\caption{Classification result of k-NN with $k=3$.
}\label{fig:knn_k=3}
\end{figure}

Now we use a simple example to illustrate how labeled data is used in k-NN to predict the class labels of those unlabeled objects. Fig. \ref{fig:knndata} shows a simple two-dimensional dataset. This dataset consists of seven labeled objects belonging to two classes and two unlabeled objects. First, we set $k=1$. In this case, each unlabeled object is assigned to the same class as its nearest neighbor. Fig. \ref{fig:knn_k=1} shows the classification result of the two unlabeled objects with $k=1$. Since the nearest neighbor of the first unlabeled object, i.e., the one that is located at the left lower corner, is labeled as class 1, this object is also labeled as 1. Similarly, since the nearest neighbor of the other unlabeled object is labeled as class 2, the class label is also predicted as class 2 for this object. When $k>1$, the neighbors of an unlabeled object possibly have different class labels, and in such cases, the unlabeled object is typically assigned to the most common class among its neighbors. Fig. \ref{fig:knn_k=3} shows the classification result of k-NN with $k=3$. It is seen that the object in the left lower corner is still labeled as class 1 as all its three nearest neighbors are in this class. However, the other object is now labeled as class 1 as two of its nearest neighbors belong to class 1, although its most nearest neighbor belongs to class 2. Here, once $k$ is decided, all the neighbors are considered to be equally important in deciding the class of the unlabeled object. Another way is to assign different weights to the neighbors so that the $k$ neighbors have different levels of significance of their votes.

%\begin{figure}[ht]
%\centerline{
%  \subfigure[]
%     {\epsfig{figure=knn1.eps,width=2in}\label{ra_fig2a}}
%  \hspace*{4pt}
%  \subfigure[Optional subcaption]
%     {\epsfig{figure=knn3.eps,width=2in}\label{ra_fig2b}}
%}
%\caption{Common caption here.} \label{ra_fig2} % common label
%\end{figure}

\subsection{Support Vector Machine}
\begin{figure}[!tpb]%figure1
\centering
\includegraphics[width=0.5\textwidth]{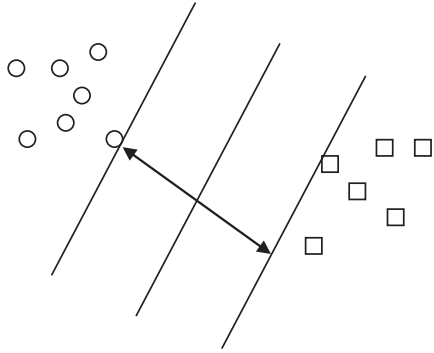}
\caption{Example of linearly separable dataset in a two dimensional space. An optimal hyperplane is the one that maximizes the distance between two classes.
}\label{fig:linearsvm}
\end{figure}

The classic Support Vector Machine (SVM) is a linear binary classifier. Given a $p$-dimensional dataset where the training samples belong to two classes, the goal of a linear classifier is to find a $p-1$ dimensional hyperplane which separates the samples in the two classes as illustrated in Fig. \ref{fig:linearsvm}. Among many of such kind of hyperplanes, the one maximizes the separation or margin of the two classes is of most interest, and the corresponding classifier is called the maximum margin classifier. In SVM, the margin is the distance from the hyperplane to the nearest samples in each of the classes. Samples located on the boundary of each class are called support vectors.

%In the training process, a model or function is learned by which the training samples are mapped into a space so that samples with different classes are separated as wide as possible.
\subsubsection{Linear SVM}
Now we formally define the linear SVM.
For a set of $n$ training samples, where each object with label $-1$ or $1$ is a $p$-dimensional vector, we may represent it as $D=\{(\textbf{x}_i,y_i)|\textbf{x}_i \in R^p, y_i \in \{-1,1\}\}^n_{i=1}$, where $X$ represents the data and $Y$ represents the label information.
 Assume that the dataset is linearly separable, then there exist $\mathbf{w}$ and $\mathbf{b}$ such that the inequalities are valid for all $\mathbf{x}_i \in D$:
\begin{align}
\mathbf{w}\cdot \mathbf{x}_i-b\geq1 \ \ \text{if} \ \ y_i=1\\
\mathbf{w}\cdot \mathbf{x}_i-b\leq-1 \ \ \text{if} \ \ y_i=-1
\end{align}
The above two equations can be written into one as below
\begin{align}
y_i(\mathbf{w}\cdot \mathbf{x}_i-b)\geq 1 \label{eq:svmconstrain}
\end{align}
Among the training samples, vectors $\mathbf{x}_i$ for which
\begin{align}
y_i(\mathbf{w}\cdot \mathbf{x}_i-b)= 1
\end{align}
are called support vectors, which define the boundary of the two classes.

The distance or margin between the two classes is $\frac{2}{\|\mathbf{w}\|}$. The goal is to find the optimal hyperplane or to decide $\mathbf{w}$ and $\mathbf{b}$ to maximize this margin subject to (\ref{eq:svmconstrain}), which requires all the training samples  to be correctly classified.
Since maximizing $\frac{2}{\|\mathbf{w}\|}$ is equivalent to minimizing $\frac{1}{2}\|\mathbf{w}\|^2$, we can solve the above maximization problem by solving the equivalent minimization problem as below

\begin{align}
\min \frac{1}{2}\|\mathbf{w}\|^2
\end{align}
subject to
\begin{align}
y_i(\mathbf{w}\cdot \mathbf{x_i}-b)\geq 1 \ \  \text{for}\ \ i=1,2,\ldots, n.
\end{align}

This constrained optimization problem can be solved with the method of Lagrange. By introducing Lagrange multipliers $\alpha_i$, the Lagrangian is constructed as

\begin{align}
\frac{1}{2}\|\mathbf{w}\|^2-\sum^n_{i=1}\alpha_i(y_i(\mathbf{w}\cdot \mathbf{x_i}-b)-1) \label{eq:svm_lagrange}
\end{align}
which can be solved by standard quadratic programming techniques.
According to the Karush-Kuhn-Tucker conditions, the solution of $\mathbf{w}$ is in the form as below:
\begin{align}
\mathbf{w}=\sum^n_{i=1}\alpha_iy_i\mathbf{x}_i \label{eq:svm_w}
 \end{align}
 The above formula shows that $\mathbf{w}$ is a linear combination of the training samples. When $y_i(\mathbf{w}\cdot \mathbf{x}_i-b)= 1$, $\alpha_i>0$; for other cases, $\alpha_i=0$. This means that $\mathbf{w}$ is only defined by a small number of support vectors, i.e., the training samples located at the boundary of the classes, rather than all the training samples.

In the above formulation, we assume that the dataset is linearly separable, or there exist a hyperplane that can divide the samples according to their class labels without any classification error. In cases that such kind of hyperplane does not exist, we may want to find a hyperplane that correctly divide the samples as many as possible. This is called the soft margin method. Slack variables $\xi_i\geq0$ are introduced to formulate this idea. The constraints are now become

\begin{align}
y_i(\mathbf{w}\cdot \mathbf{x}_i-b)\geq 1-\xi_i
\end{align}
Since a larger $\xi_i$ corresponds to a larger error in the classification of $x_i$, we want to penalize large $\xi_i$ through minimizing the objective function as below
\begin{align}
\min \frac{1}{2}\|\mathbf{w}\|^2+C\sum^n_{i=1}\xi_i
\end{align}
where $C$ is the weight parameter of the penalty term.
With Lagrange multipliers $\alpha_i\geq0$ and $\beta_i\geq0$, the problem to be solved is written as
\begin{align}
\min \frac{1}{2}\|\mathbf{w}\|^2+C\sum^n_{i=1}\xi_i-\sum^n_{i=1}\alpha_i(y_i(\mathbf{w}\cdot \mathbf{x_i}-b)-1+\xi_i)-\sum^n_{i=1}\beta_i\xi_i
\end{align}

\subsubsection{Kernel SVM}
In many cases, the data is not linearly separable. As illustrated in Fig. \ref{fig:kernel}, mapping the original space into a high or infinity dimensional feature space, i.e., $\mathbf{x}\rightarrow\phi(\mathbf{x})$, possibly makes the data easier to be separated. Kernel-based approach use a kernel function $\kappa$ to calculate the inner product of the vectors in the high dimensional space in terms of the vectors in the original space:
\begin{figure}[!tpb]%figure1
\centering
\includegraphics[width=0.7\textwidth]{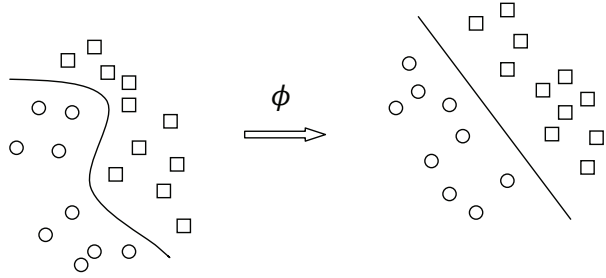}
\caption{ A non-linearly separable dataset becomes linearly separable after mapping $\phi$
}\label{fig:kernel}
\end{figure}
\begin{align}
\kappa(x_i,x_j)=\phi(\mathbf{x}_i) \cdot \phi(\mathbf{x}_j)=\phi(\mathbf{x}_i)^T\phi(\mathbf{x}_j)\label{eq:kernel}
\end{align}

 As $\|\mathbf{w}\|=\mathbf{w}^T\mathbf{w}$, substituting (\ref{eq:svm_w}) into (\ref{eq:svm_lagrange}), the dual of SVM is the following optimization problem:
\begin{align}
\max_{\alpha_i} = \sum^n_{i=1}\alpha_i -\frac{1}{2} \sum^n_{i=1}\sum^n_{j=1}\alpha_i\alpha_j y_i y_j x^T_ix_j
\end{align}
subject to
\begin{align}
\alpha_i \geq 0 \ \  \text{for}\ \ i=1,2,\ldots, n. \\
\sum^n_{i=1}\alpha_iy_i=0
\end{align}

By substitute $\mathbf{x}_i$ with $\phi(\mathbf{x_i})$ in the above formula, we get the objective function in the mapped space, and with the kernel function given in (\ref{eq:kernel}), we have the following form without defining the mapping explicitly:
\begin{align}
\max_{\alpha_i} = \sum^n_{i=1}\alpha_i -\frac{1}{2} \sum^n_{i=1}\sum^n_{j=1}\alpha_i\alpha_j y_i y_j \kappa(\mathbf{x}_i,\mathbf{x}_j)
\end{align}
Below are the three commonly used kernels:
\begin{itemize}
\item Polynomial kernel
\begin{align}
\kappa(\mathbf{x}_i,\mathbf{x}_j)=(\mathbf{x}_i\cdot\mathbf{x}_j+1)^d
\end{align}
\item Gaussian kernel
\begin{align}
\kappa(\mathbf{x}_i,\mathbf{x}_j)=e^{-\frac{\|\mathbf{x}_i-\mathbf{x}_j\|^2}{\beta}}\label{eq:gaussiankernel}
\end{align}
\item Hyperbolic tangent kernel
\begin{align}
\kappa(\mathbf{x}_i,\mathbf{x}_j)=tanh(h\mathbf{x}_i\cdot \mathbf{x}_j+c)
\end{align}
where $h$ is the scale factor and $c$ is the offset.
\end{itemize}
Since any positive-definite matrix could be treated as a kernel matrix, kernel SVM can be used to make prediction based on a similarity matrix, which records pairwise similarities between objects. To make sure kernel SVM performs stably, some preprocess is needed if the similarity matrix given is not positive-definite.

\subsection{Bayesian classification}
Bayesian classifiers are statistical classifiers based on Bayes theorem. A Bayesian classifier generates the probability or membership of an object with respect to each of the classes. Assume $X$ is an object that is to be classified or labeled and $Y$ is the hypothesis that $X$ belongs to some class, then $P(Y=c/X)$ is the probability that $X$ belongs to the $c$th class. According to the Bayes theorem, this posterior probability of $Y=c$ conditioned on $X$ can be calculated with posterior probability $P(X/Y=c)$, and prior probabilities $P(X)$ and $P(Y)$:
\begin{align}
P(Y=c/X)=\frac{P(X/Y=c)P(Y=c)}{P(X)}
\end{align}
In the above formula, $P(X)$ is constant for any $c$. If $P(Y=c)$ is unknown, it is usually assumed that all classes have equal probability or it is estimated by $\frac{n_c}{n}$, the ratio of the number of objects in class $c$. The left problem is how to calculate $P(X/Y=c)$. To simplify computation, the values of attributes are assumed to be conditionally independent to each other, i.e., given the class label of an object, there are no dependence relationships among the attributes. Assume $x_j$ is the value of the $j$th feature, and there are $p$ features in total, then based on this assumption,
\begin{align}
P(X/Y=c)=\prod^p_{j=1}P(x_j/Y=c)
\end{align}
and the classifier is called the Naive Bayes Classifier.

If the $k$th attribute is categorical, then
\begin{align}
P(x_j/Y=c)=\frac{n_{jc}}{n_c}
\end{align}
where $n_{c}$ is the number of objects in class $c$, and $n_{jc}$ is the number of objects in class $c$ that have the value of the $k$th attribute equal to $x_j$.

If the $k$th attribute is continuous-values with a probability distribution $g$, e.g., the Gaussian distribution with mean $\mu_c$ and standard deviation $\delta_c$, then

\begin{align}
P(x_j/Y=c)=g(x_j)=\frac{1}{\sqrt{2\pi}\delta_c}e^{-\frac{(x_j-\mu_c)^2}{2\delta^2_c}}
\end{align}

Once the posterior probabilities $P(X/Y=c)$ for all $c=1,2,\ldots,k$ are calculated, $X$ is assigned to the class with the largest posterior probability, i.e., $X$ is labeled as class $f$, where $f=\arg\max_cP(X/Y=c)$.

\subsection{Decision Trees}
A decision tree is a tree structure where each internal node denotes a test on an attribute, each branch denotes an outcome of the test, and each leaf node represents a class. Once a decision tree has been constructed with training data, a new sample is tested against the decision tree from the top node to the leaf node which corresponds to the predicted class of the new sample.

Given a set of training objects, a decision tree is built in a top-down recursive divide and conquer manner. A critical problem need to be considered in construction of the tree is how to select the attributes for testing. Entropy or equivalently information gain and Gini index are commonly used for attribute selection.
The entropy measures the purity of the partitions, the smaller the entropy or the larger the information gain, the purer the partitions are. Thus, the attribute with the minimum entropy or highest information gain is chosen as the test attribute for the current node.
Assume the training data consists of $n$ labeled objects are distributed in $k$ classes, each class contains $n_c$ objects, then the expected information needed to classify a given sample is
\begin{align}
E=-\sum^k_{c=1}p_clog_2p_c
\end{align}
where $p_c$ is the probability that an arbitrary object belongs to class $c$. It is estimated by $\frac{n_c}{n}$.
For a feature $a$, which has $h$ distinct values, the entropy based on the partitioning into $k$ subsets by $a$ is calculated by
\begin{align}
E(a)=\sum^h_{j=1} P(j)E(j)
\end{align}
where
$P(j)=\frac{n_j}{n}$, $n_j$ is the number of objects of which the value of feature $a$ is equal to $j$, and
\begin{align}
E(j)=-\sum^k_{c=1} p_{cj}log_2p_{cj}
\end{align}
is the entropy of the $j$th value of the $a$th attribute, $p_{cj}=\frac{n_{cj}}{n_j}$
\begin{align}
G(a)=E-E(a)
\end{align}

\begin{table}
\centering
\tbl{Weather Data}
{\begin{tabular}{llllll}\hline
id &Outlook   &   Temperature  &  Humidity   &  Windy   &  Play   \\\hline
1&Sunny     &  Hot           & High        & False    & No     \\
2&Sunny     &   Hot          &  High       &  True    &  No      \\
3&Overcast  &   Hot          &  High       &  False   &  Yes     \\
4&Rainy     &   Mild         &  High       &  False    & Yes     \\
5&Rainy     &   Cool         &  Normal     &  False    & Yes     \\
6&Rainy     &   Cool         &  Normal     &  True     & No      \\
7&Overcast  &   Cool         &  Normal     &  True     & Yes     \\
8&Sunny     &   Mild         &  High        & False    & No      \\
9&Sunny     &   Cool         &  Normal      & False    & Yes     \\
10&Rainy     &   Mild         &  Normal      & False    & Yes     \\
11&Sunny     &   Mild         &  Normal      & True     & Yes     \\
12&Overcast    & Mild& High       &  True    &  Yes     \\
13&Overcast   &  Hot           & Normal     &  False   &  Yes     \\
14&Rainy      &  Mild          & High       &  True    &  No      \\\hline
\end{tabular}}\label{tab:weatherdata}
\end{table}

Now we use the Weather data in Table \ref{tab:weatherdata} as an example to show how to calculate the Entropy of each attribute. This data consists of fourteen samples described by four attributes: Outlook, Temperature, Humidity and Windy. These fourteen samples belong to two classes: \emph{Play} or \emph{Not-Play}. It is shown that
$P(Play=Yes)=\frac{9}{14}$, $P(Play=No)=\frac{5}{14}$, So the Entropy of Play or the expected information needed to classify a sample is
\begin{align}
E=-(\frac{9}{14}log_2\frac{9}{14}+\frac{5}{14}log_2\frac{5}{14})=0.940
\end{align}
Now we calculate the Entropy of the attribute Outlook. This attrbute has three values \emph{Sunny}, \emph{Overcast}, and \emph{Rainy}, which occurs 5, 4, and 5 times, respectively, i.e.,
$P(Sunny)=\frac{5}{14}$, $P(Overcast)=\frac{4}{14}$ and $P(Rainy)=\frac{5}{14}$.
Among the five samples of which Outlook is Sunny, two are \emph{Play}, three are \emph{Not-Play}, thus the Entropy of Sunny is
\begin{align}
E(Sunny)=-(\frac{2}{5}log_2\frac{2}{5}+\frac{3}{5}log_2\frac{3}{5})=0.971
\end{align}
Similarly
\begin{align}
E(Overcast)=-(\frac{4}{4}log_2\frac{4}{4}+0log_20)=0\\
E(Rainy)=-(\frac{3}{5}log_2\frac{3}{5}+\frac{2}{5}log_2\frac{2}{5})=0.971\\
\end{align}
So the Entropy of \emph{Outlook} is
\begin{align}
&E(Outlook)\\
&=P(Sunny)E(Sunny)+P(Overcast)E(Overcast)+P(Rainy)E(Rainy)\\
&=\frac{5}{14}0.971+\frac{4}{14}0+\frac{5}{14}0.971=0.694
\end{align}
and the Information Gain of \emph{Outlook} is
\begin{align}
Gain(Outlook)=E-E(Outlook)=0.940-0.694=0.246
\end{align}
With the same steps, we can calculate the Gain of the other three attributes: $Gain(Temperature) = 0.029$, $Gain(Humidity) = 0.152$, and $Gain(Windy) = 0.048$. Since Outlook has the largest Gain, it is the best attribute of the current stage that should be selected for testing.

\begin{figure}[!tpb]%figure1
\centering
\includegraphics[width=0.7\textwidth]{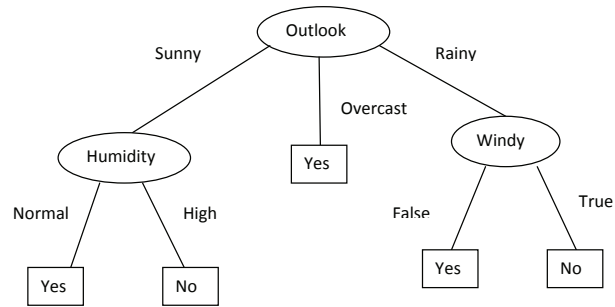}
\caption{The decision tree of the Weather data.
}\label{fig:tree_weather}
\end{figure}

Once the best attribute is decided and represented as an intermediate node of the tree, branches below this node are added where each branch corresponds to a possible value this attribute takes. For each value, take the subset of samples having this value of the current attribute as the input of the next iteration for further splitting. This process continues until all samples under consideration have the same class label. A complete decision tree of the Weather data is shown in Fig. \ref{fig:tree_weather}.

\begin{figure}[!tpb]%figure1
\centering
\subfigure{\includegraphics[width=0.46\textwidth]{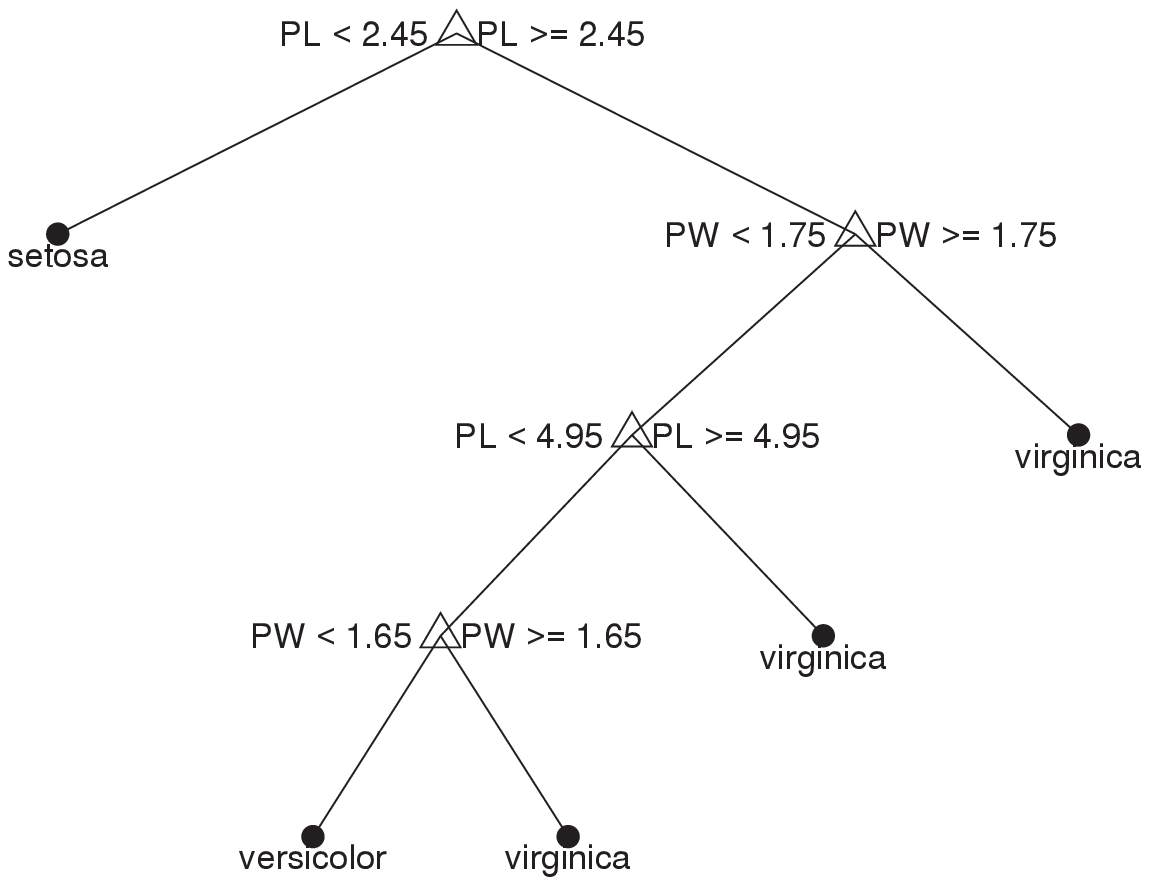}}\hfill
\subfigure{\includegraphics[width=0.46\textwidth]{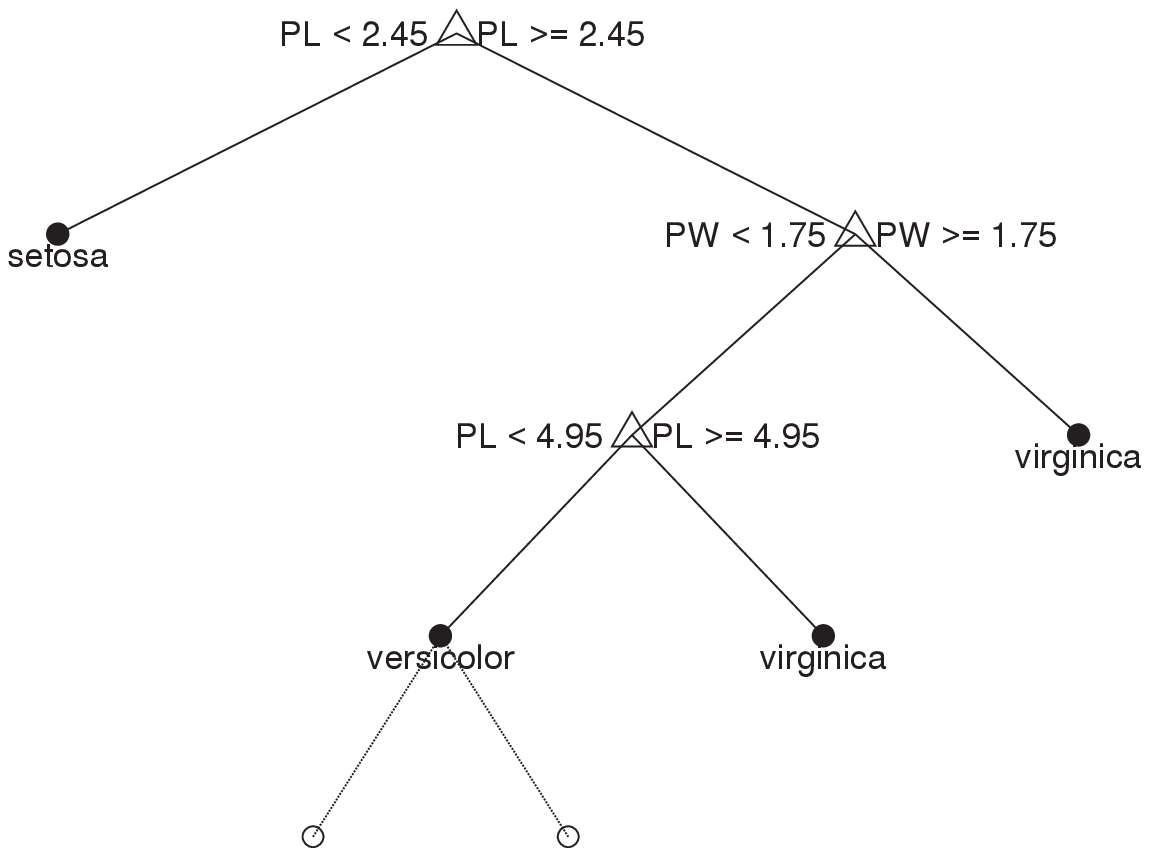}}
\caption{The decision tree of the Iris data. (a) The unpruned three, (b) The tree with pruning.
}\label{fig:Iris}
\end{figure}

The tree constructed to correctly classify all the training samples may be over-fitting. Pruning handle the over-fitting problem by removing least reliable branches. Other than a higher classification accuracy, pruning also results in a simplified tree which makes the test process faster. Pruning performed during the construction of the tree is called Prepruning. It stops the construction early with less purity. Pruning can also be performed by removing branches from a fully grown tree. This type is called the post-pruning. Fig. \ref{fig:Iris} shows the unpruned and pruned decision three of the Fisher's iris data. This dataset consists of 50 samples from each of three species of Iris (setosa, virginica and versicolor). Four features were measured from each sample: sepal length (SL), sepal width (SW), petal length (PL), and petal width (PW).

ID3 is a popular decision tree algorithm proposed by Ross Quinlan \cite{quinlanID3}, and C4.5 \cite{quinlanC45} is an extension of ID3 with improved computing efficiency, and other more functions, including dealing with continuous values, handling attributes with missing values, and avoiding over fitting. Another algorithm called Classification and regression trees (CART) proposed by Leo Breiman \cite{Breiman84CART} produces either classification or regression binary trees, depending on whether the dependent variable is categorical or numeric, respectively. The study in \cite{chen11classiftree} reviews tree-based classification approaches and their applications in bioinformatics.

\subsection{Regression models for classification}
Other than these previously reviewed supervised learning methods which are widely used for classification, regression models may also be used for classification analysis. Regression methods the relationship between a dependent variable and one or more independent variables. Specifically, regression is to analyse how the value of the dependent variable changes when any one of the independent variable varies while other independent variables fixed. The dependent variable is the output variable or response variable, and the independent variables are input variables or explanatory variables. Next we discuss two regression models namely the Logistic Regression and Regularized Least Squares, which are frequently used for classification purpose.
\subsubsection{Logistic Regression}
Logistic Regression is a type of binomial regression that predicts the probability of the outcome of a ``yes or no'' type trial using logistic function. Formally, the Logistic Regression models the relation between dependent variable $y_i$ and independent variables $\textbf{x}_i=(x_{i1} \ldots x_{ip})^T$ by

\begin{align}
y_i=\frac{1}{e^{-(\mathbf{x}_i^T\boldsymbol{\beta}+\epsilon_i)}+1}
\end{align}
where $\boldsymbol{\beta}=(\beta_1 \ldots \beta_p)^T$ are regression coefficients, and $\epsilon_i$ is the error term. Let
\begin{align}
t=\sum^p_{j=1}\beta_jx_{ij}+\epsilon_i=\mathbf{x}_i^T\boldsymbol{\beta}+\epsilon_i \ \ for \ \ i=1,2,\ldots n
\end{align}
then $y_i=f(t)$, where $f(t)$ is the logistic function
\begin{align}
f(t)=\frac{1}{e^{-t}+1}
\end{align}
A property of the logistic function is like distribution function, its output is between 0 and 1 for any input in the full range from negative infinity to positive infinity, i.e., $f(t) \in [0,1]  \ \ \text{for}\ \ t \in (-\infty, \infty)$.
The coefficients are usually estimated with maximum likelihood estimation with iterative algorithms such as Newton's Method.
Once the coefficients are learned, the logistic regression can be used for binary classification where the predicted value $\hat{y}_i$ is the probability of being ``yes''.
\subsubsection{Regularized Least Squares}
Unlike many other regression models, such as Logistic Regression, the Regularized Least Squares (RLS) method does not require the examples to be represented as feature vectors explicitly as it learns the model and makes prediction with a kernel matrix $\mathbf{K}$, where each entry $k_{ij} \in K=\kappa(\mathbf{x}_i,\mathbf{x}_j)$ is defined by a certain kernel function, e.g., Gaussian kernel in (\ref{eq:gaussiankernel}).
For a dataset with labels $\mathbf{y}=(y_1  y_2 \ldots y_n)^T$, and kernel matrix $\mathbf{K}$, the Regularized Least Squares (RLS) is to find coefficients $\mathbf{c}=(c_1 c_2 \ldots c_n)^T$ to minimize the following value
\begin{align}
 \frac{1}{2}\Vert \mathbf{y}-\mathbf{K}\mathbf{c}\Vert^2_2+\frac{\delta}{2}\mathbf{c}^T \mathbf{K} \mathbf{c}
\end{align}
where the first term is the least squares term and the second term is the regularization term with weight $\delta$. The solution of $\mathbf{c}$ that minimizes the above value has a simple closed form as below
\begin{align}
\mathbf{c}=(\mathbf{K}+\delta I)^{-1}\mathbf{y}
\end{align}
Once $\mathbf{c}$ is obtained, we can use it to predict the label $\hat{y}$ of a new data object $\mathbf{\hat{x}}$ by
\begin{align}
\hat{y}=\mathbf{\hat{k}}^T(\mathbf{K}+\delta \mathbf{I})^{-1}\mathbf{y}
\end{align}
 $\mathbf{\hat{k}}$ is an $n$-dimensional vector where each dimension $\hat{k_i}$ is the value of the kernel function between this object and a training example, i.e., $\hat{k_i}=\kappa(\mathbf{\hat{x}},\mathbf{x_i})$.

 In real applications, the similarity matrix recording a certain type of similarity between each pair of examples may be treated as a kernel matrix. Since kernel matrix is positive definite, some preprocessing may be needed to transform the given similarity matrix into a positive definite matrix.
\subsection{Ensemble classifier}
\begin{figure}[!tpb]%figure1
\centering
\includegraphics[width=0.7\textwidth]{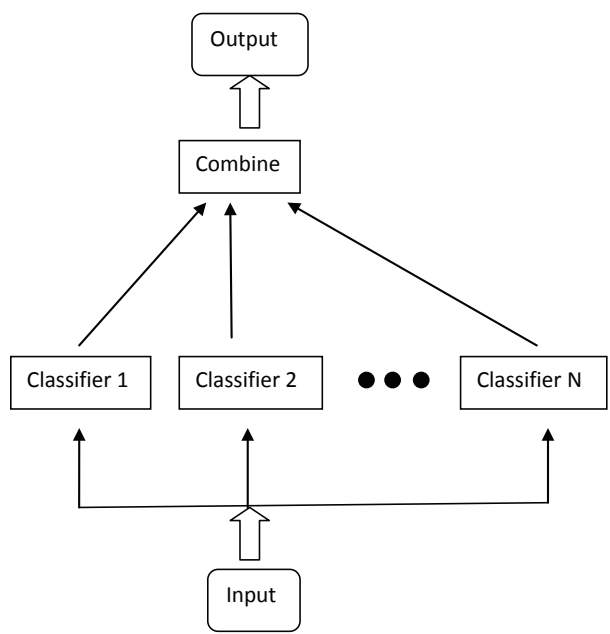}
\caption{ An ensemble of classifiers.
}\label{fig:ensemble}
\end{figure}

An ensemble classifier is not a specific type of classifier as those introduced earlier. Instead, it is a classifier ensemble, which combines or aggregates the predictions of several individually trained classifiers called base classifiers to produce a final result. A simple enselble classifier is illustrated in \ref{fig:ensemble}. Through aggregating, the prediction of an ensemble classifier is usually more accurate than any of the individual classifiers. An important problem is how to train each of the base classifiers. Since ensemble makes sense only if the outputs of the base classifiers are different. To generate disagreements in the prediction, base classifiers may be trained with different initial weights, different parameters, different subsets of features, and different portions of training set. The two well known ensemble methods: Bagging \cite{Breiman96bagging}, and Boosting \cite{Schapire1990boost,Freund1997} mainly focus on the last way to train the base classifiers, and the other well known method Random Forest \cite{Breiman01radomforest} makes use of the last two ways.

In the Bagging method, each classifier is trained on a random sample of the training set. More specifically, a set of sample to be used for training a base classifier is generated by randomly drawing with replacement from the training samples. Although each individual classifier could result in higher test-set error when trained with a subset of training samples, the combination of them can produce lower test-set error than using the single classifier trained with all the training samples. \cite{Breiman96bagging} showed that Bagging is effective on ``unstable'' learning algorithms, such as decision tree and neural network, where small changes in the training set result in large changes in predictions.
Unlike Bagging, where the generation of training set for one classifier is  independent on other classifiers, in Boosting \cite{Breiman1996arcing,Freund1997}, the training set used for each base classifier is chosen based on the performance of the earlier classifiers. Examples that are incorrectly predicted by previous classifiers are selected more often than those were correctly predicted. Doing this, Boosting attempts to make subsequent classifiers be better able to predict examples for which the current ensemble's performance is poor.
The Random Forest \cite{Breiman01radomforest} combines the Bagging idea to select training samples and random selection of features. The selection of a random subset of features is an example of the random subspace method \cite{Ho98subspace}, which is especially useful for handling high-dimensional data, e.g., gene expression data. Projecting the original high dimensional space into different low subspaces so that the problems caused by high-dimensionality are avoid. Although decision tree is often used as base classifiers in these ensemble methods, other types of classifiers may also be used to produce base predictions in an ensemble.

Once all the base classifiers are trained, they generate predictions for new samples to be classified. Voting is a commonly used way to combine these predictions to give the final class label for the input. Assuming that the majority of the classifiers would make the correct prediction, voting labels the sample as the class that predicted by most of the base classifiers. Instead of equally weighting the classifiers, the aggregating weights for each base classifier may also be adapted according to their performance \cite{Freund1997}.

\section{Drug-target interaction prediction}
In this section, we take the drug-target interaction prediction as an example to present detailed discussion on how classification is used to handle a specific  task in biology.
   Some background knowledge of drug-target interaction prediction is first given. After that, recently studies on using classification for drug-target interaction prediction are discussed. Finally, experimental studies on benchmark datasets are given to evaluate the performance of several different classification approaches in drug-target interaction prediction.

\subsection{Background}
Identification of drug-target interaction is an important part of the drug discovery pipeline. The great advances in molecular medicine and the human genome project provide more opportunities to discover unknown associations in the drug-target interaction network. These new interactions may lead to the discovery of new drugs and also are useful for helping understand the causes of side effects of existing drugs. Since experimental way to determine drug-target interactions is costly and time-consuming, \emph{in silico} prediction comes out to be a potential complement that provides useful information in an efficient way.

Traditional approaches for this task are generally categorized into drug-based approaches and target-based approaches. Drug-based approaches screen candidate drugs, compounds or ligands to predict whether they interact with a given target based on the assumption that similar drugs share the same target. The similarity of two drugs are measured in different ways with respect to different aspects. Other than comparing drugs according to their chemical structures \cite{Martin02}, side-effect has also been used to measure the similarity between drugs \cite{Campillos08}. Assuming that similar targets bind to the same ligand, target-based approaches, on the other hand, compare proteins to predict whether they bind to the given ligand, or whether they are the targets of the given drug or compound. More specifically, for a given drug, new targets are identified by comparing candidate proteins to the known targets of this drug with respect to certain descriptors such as amino acid sequence, binding sites, or ligands that bind to them. The authors of \cite{Haupt11} review computational methods to find new targets for already approved drugs for the treatment of new diseases based on the structural similarity of their binding sites. Candidtae targets are compared by the chemical similarity of ligands that bind to them \cite{Keiser09}. Different from these classic drug-based or target-based approaches, chemogenomics approaches have been proposed to consider the interactions between drugs and a protein family rather than a single target \cite{Caron01,Kubinyi04,Rognan07,Jacob08}.

 Recently, machine learning approaches have been applied to this task to explore the whole interaction space. In the supervised bipartite graph learning approach \cite{Yamanishi08}, the chemical space and the geometric space are mapped into a unified space so that those interacted drugs and targets are close to each other while those non-interacted drugs and targets are far away from each each other. After the mapping function to such a unified space is learned, the query pair of drug and target are also mapped in the same way to that unified space, and the probability of interaction between them is the closeness that they are in the mapped space. It has been shown that the combination of supervised learning independently based on
drug and target performs very well \cite{Bleakley09}. This approach is called the Bipartite Local Model (BLM). For a query pair of drug and target, a model of the query drug is learned with a certain classifier based on the information of its known targets. Then the probability of interaction between this drug and the query target is predicted with this model. The same procedure is applied to obtain the probability of interaction between them from the target side. Finally, an overall probability of interaction for the query pair is calculated by combing these two probabilities. It has been reported that the result based the knowledge of both directions, i.e., from the drug side and from the target side, is much better than those based on each single one.
The same idea is adopted by another two following work. Semi-supervised approach is used instead of supervised approach to learn the local model \cite{Xia10}. Laarhoven found that only use the kernel based on the topology of the known interaction network is able to obtain a very good performance, although together with other types of similarities can further improve the results \cite{Laarhoven11}. Other than using one type of drug-drug similarity and one type of target-target similarity, \cite{Perlman11} use multiple types of drug-drug similarities and target-target similarities and combine them as features to describe each drug-target pair to learn the logistic regression model. Next, we present the details of how the drug-targe prediction task is handled by three types of classification problems.

\subsection{A binary classification problem}
A relatively stratforward way to predict whether a given pair of drug-target interacts is to model it as a binary classification problem as in Ref.~\cite{Perlman11}. The key problem is how to extract a set of features based on different biological sources to charactorize or represent each drug-target pair. This has been done in three steps in Ref.~\cite{Perlman11}. First, five drug-drug similarities and three gene-gene similarities are calculated based on different bilological and chemical sources. Then, the drug and gene similarity measures are combined as features to describe each drug-target pair. Feature selection is performed to select important features. Finally, the classifier is trained with the labeled samped decribed with selected features. In this study, Logistic regression is used for classification.

\begin{figure}[!tpb]%figure1
\centering
\includegraphics[width=0.8\textwidth]{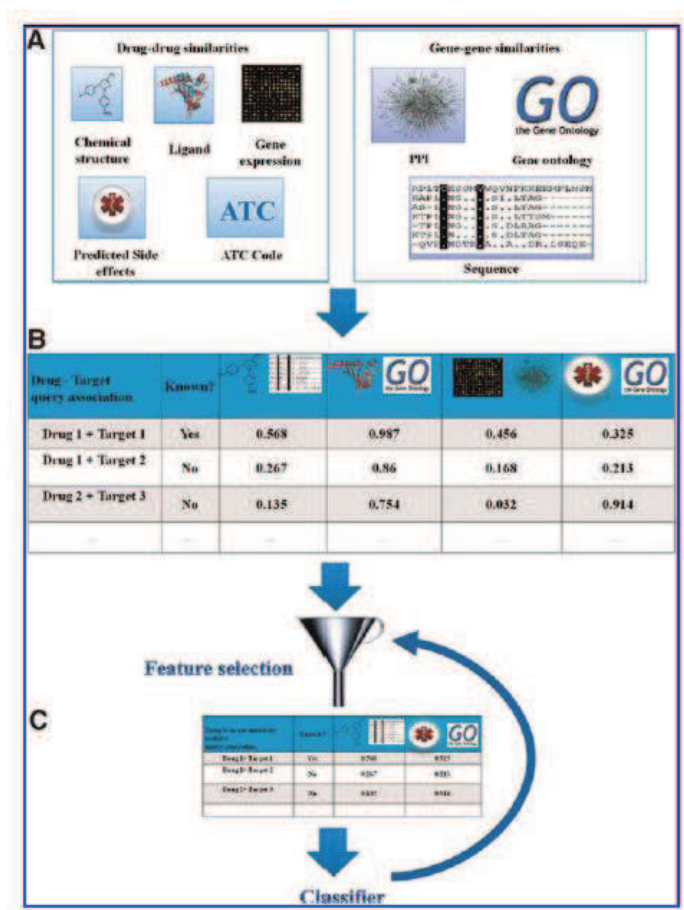}
\caption{Algorithm pipeline. (A) comprised
of formation of drug-drug
and gene-gene similarity matrices, (B) integration of the similarities to
classification features, (C) classification
with feature selection.
}\label{fig:perl}
\end{figure}

The whole process is shown in Fig. \ref{fig:perl} \cite{Perlman11}. The drug-drug similarity measures were computed using chemical strucute, Ligand, drug side effects, drug response gene expression profiles, and the
Anatomical, Therapeutic and Chemical (ATC) classification system code. The gene-gene similarity measures
used are based on protein-protein interactions, sequence, and Gene Ontology (GO). Once all these drug-drug similarities and target-target similarities are obtained, each feature is constructed based on one drug-drug similarity and one target-target similarity. Specifically, calculated by
combining the drug-drug similarities between the query drug and other drugs and the gene-gene similarities
between the query gene and other target genes across all true drug-target associations. Therefore, fifteen features are constructed in such a way. After feature selection, ten features are finally selected. Table \ref{Tab:feature} shows the results in terms of AUC (area under ROC curve) and AUPR (area under precision-recall curve) with all the features, all the selected features and  each single selected feature. Here AUC and AUPR are two performance evaluation measures. It is shown that using ten selected feature gives a comparable result with all the fifteen features, which is much better than using any of a single feature. It is also shown that when used indivudually, the combination of Ligand and sequence similarity gives the best feature. Once each drug-target pair is represented as a vector of these feaures, the prediction problem of whether a query pair interacts simply becomes a binary classification problem that can be solved by many existing classification algorithms, e.g., the Logistic regression as used in this paper. Other than develping a good data presentation through aggregation of multiple data sources, some other studies focus more on design of new learning algorithms. Next we introduce two recently proposed learning algorithms, namely the Bipartite Graph Learning  and the Bipartite Local Model.

\begin{table}[!t]
\tbl{Comparison of AUC and AUPR for the four datasets}
{\begin{tabular}{lllll}\hline
&All features &0.905 &0.935\\
&selected features &0.908 &0.935\\
Ligand &Sequence similarity &0.851 &0.867\\
Ligand &GO semantic similarity &0.845 &0.867\\
Predicted Side Effect &GO semantic similarity &0.832 &0.863\\
ATC similarity &GO semantic similarity &0.81 &0.858\\
Ligand &PPI closeness &0.809 &0.844\\
Chemical &GO semantic similarity &0.805 &0.84\\
ATC similarity &PPI closeness &0.762 &0.809\\
Chemical &Sequence similarity &0.749 &0.763\\
Predicted Side Effect &PPI closeness &0.729 &0.759\\
Co-expression &Sequence similarity &0.724 &0.748\\\hline
\end{tabular}}\label{Tab:feature}
\end{table}
\subsection{Bipartite graph learning (BGM)}
We assume that the problem under consideration is to predict new interactions between $n_d$ drugs and $n_t$ targets. An $n_d \times n_t$ matrix $\mathbf{A}$ is used to record these known interactions, i.e., $a_{ij} \in \mathbf{A}=1$ if the $i$th drug denoted as $d_i$, is known to interact with the $j$th target denoted as $t_j$. All other entries of $\mathbf{A}$ are 0. Assume $n_i$ interactions in total involves $m_d$ drugs and $m_t$ targets and $m_d<n_d$ and $m_t<n_t$. This means there are some new drug and target candidates and the corresponding rows and columns of $\mathbf{A}$ are all 0. Other than the interaction network, $\mathbf{S}_d$ and $\mathbf{S}_t$ are the chemical similarity matrix of drug and the sequence similarity matrix of target, respectively.

\begin{figure}[!tpb]%figure1
\centering
\includegraphics[width=0.9\textwidth]{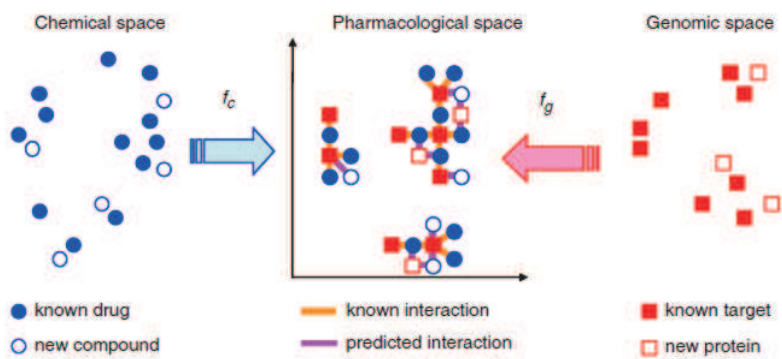}
\caption{Bipartite graph model.
}\label{fig:BGM}
\end{figure}

The bipartite graph learning method learns the correlation between the chemical/genomic space and the interaction space, which is called the `pharmacological space'. As illustrated in Fig. \ref{fig:BGM} \cite{Yamanishi08}, first, the compounds and proteins are embedded into a unified space called `pharmacological space'. The mapping function or model between the chemical/genomic space and the pharmacological space is learned. With this model, any query pair of compounds and proteins are mapped onto the same pharmacological space. The compound-protein pairs under testing are predicted to be interacting if the two are closer than a threshold in the pharmacological space. The whole process consists of the following steps \cite{Yamanishi08}:

\begin{itemize}

\item Step 1: construct a graph-based similarity matrix
\begin{align} \mathbf{K}=\left( \begin{array}{cc}
\mathbf{K}_{cc} & \mathbf{K}_{cg} \\
\mathbf{K}_{cg}^T & \mathbf{K}_{gg}
 \end{array} \right)\end{align}
where the entries of each matrices are calculated as
\begin{align}
\mathbf{K}_{cc}=exp(-\frac{d^2_{c_ic_j}}{h^2})\\
\mathbf{K}_{gg}=exp(-\frac{d^2_{g_ig_j}}{h^2})\\
\mathbf{K}_{cg}=exp(-\frac{d^2_{c_ig_j}}{h^2})\\
\end{align}
where $d$ is the shortest distance between two objects (compounds or proteins) on the bipartite graph. The symmetric matrix $\mathbf{K}$ has a scale of $(n_c+n_d) \times (n_c+n_d)$. After $\mathbf{K}$ is constructed, eigenvalue decomposition is performed to $\mathbf{K}$ to get $\mathbf{U}$:
\begin{align}
\mathbf{K}=\Gamma\Lambda^{1/2}\Lambda^{1/2}\Gamma^T=\mathbf{U}\mathbf{U}^T
\end{align}
where $\Lambda$ is the diagonal matrix with the diagonal elements the eigenvalues and the columns of matrix $\Gamma$ are the corresponding eigenvectors. Write $\mathbf{U}$ with its row vectors: $\mathbf{U}=(\mathbf{u}_{c_1},\ldots, \mathbf{u}_{cn_c},\mathbf{u}_{g_1},\ldots,\mathbf{u}_{gn_g})^T$.
\item Step 2: For $i=\{1,\ldots, {c_n}\}$ and $j=\{1, \ldots, {g_n}\}$, learn $\mathbf{w}_{ci}$ and $\mathbf{w}_{gj}$ by assuming the following relation, which is a variant of the kernel regression model:
\begin{align}
\mathbf{u}_{ci}=\sum^{n_c}_{i=1}s_c(x,x_{ci})\mathbf{w}_{ci} + \epsilon\\
\mathbf{u}_{gj}=\sum^{n_g}_{j=1}s_g(x,x_{gj})\mathbf{w}_{gj} + \epsilon\\
\end{align}

\item Step 3: mapping the query compound $c_q$ and protein $g_q$ with learned $\mathbf{W}_c$, and $\mathbf{W}_g$:
\begin{align}
\mathbf{u}_{cq}=\sum^{n_c}_{i=1}s_c(c_{q},c_i)\mathbf{w}_{ci}\\
\mathbf{u}_{gq}=\sum^{n_g}_{j=1}s_g(g_{q},g_j)\mathbf{w}_{gj}\\
\end{align}

\item Step 4: The score of interaction between $c_q$ and $g_q$ denoted as $p_{c_q,g_q}$ is calculated as the inner product of the feature vectors in the mapped space
\begin{align}
p_{c_q,g_q}=<\mathbf{u}_{cq}, \mathbf{u}_{gq}>
\end{align}
\end{itemize}

\subsection{Bipartite local model (BLM)}
\begin{figure}[]%figure1
\centering
\includegraphics[width=0.8\textwidth]{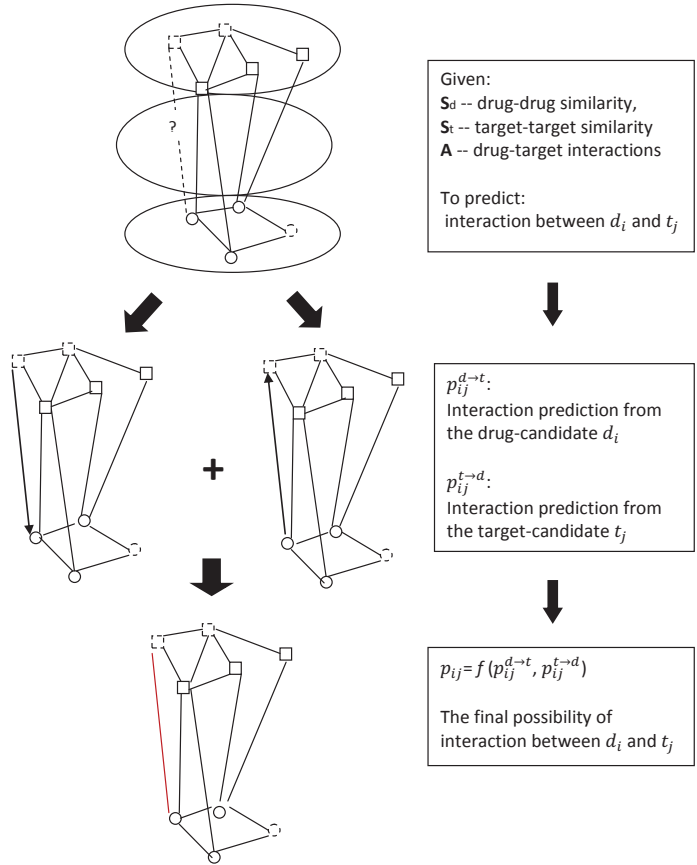}
\caption{Drug-target interaction prediction with learning from the drug and target independently.
}\label{fig:galgo}
\end{figure}

To predict $p_{ij}$, the probability that a drug $d_i$ and a target $t_j$ interacts, the basic bipartite local model is described as follows. A local model of $d_i$ is first learned based on the known targets of this drug and the similarities between these targets. This model is then used to predict $p_{ij}^{d \scriptscriptstyle \rightarrow t}$ the probability of interaction between this drug to the tested protein. The model learning and prediction process is performed independently from the query target side to get $p_{ij}^{t\scriptscriptstyle\rightarrow d}$. Once both $p_{ij}^d$ and $p_{ij}^t$ are calculated, they are combined with some function $f$ to get the final result $p_{ij}=f(p_{ij}^{d \scriptscriptstyle \rightarrow t}, p_{ij}^{t\scriptscriptstyle\rightarrow d})$. Fig. \ref{fig:galgo} \cite{Mei12} illustrates the idea of drug-target interaction prediction with learning from the drug and target independently.

This framework was first proposed in \cite{Bleakley09}, and then was further studied in Ref.~\cite{Xia10} and Ref.~\cite{Laarhoven11}. Under the same BLM framework, different results may be produced due to the differences in drug-drug similarity $\mathbf{S}_d$ and target-target similarity $\mathbf{S}_t$, the classifier, and the way how $p_{ij}^{d \scriptscriptstyle \rightarrow t}$ and $p_{ij}^{t \scriptscriptstyle \rightarrow d}$ is combined, i.e., the function $f$. For example, in \cite{Bleakley09}, Support Vector Machine (SVM) is used as the classifier using the chemical structure similarity for drug and sequence similarity for protein targets, respectively. The same types of similarity data is used in \cite{Xia10}, but with a semi-supervised approach for local model learning. In \cite{Laarhoven11}, network topology based similarity for drug and target are calculated and combined with the chemical structure similarity and sequence similarity, respectively, to give the final pairwise drug similarities and pairwise target similarities, and the Regularized Least Squares (RLS) is used for model learning. So far, simple combination functions are shown good enough to get the final prediction based on the two individually obtained ones, e.g., $p_{ij}=\max\{p^{d \scriptscriptstyle \rightarrow t}_{ij},p^{t \scriptscriptstyle \rightarrow d}_{ij}\}$ is used in \cite{Bleakley09}, and $p_{ij}=0.5(p^{d \scriptscriptstyle \rightarrow t}_{ij}+p^{t \scriptscriptstyle \rightarrow d}_{ij})$ is used in \cite{Laarhoven11}.

\subsection{Enhanced BLM with training data inferring for new drug/target candidates}
Generally, supervised learning performs better than unsupervised
learning. However, a good performance of supervised learning is largely dependent on the amount and quality of the labeled training data. When the drug candidate is new, it has no existing targets that can be used as positive labeled training data and the model for this drug thus cannot be learned. Similarly, supervised local model learning does not work for new target candidates.
To extend the application domain of BLM to new drug and target candidates, in Ref.~\cite{Mei12}, we present a training data inferring procedure and integrate it into BLM. Based on the assumption that drugs which are similar to each other interact with the same targets, training data for a new drug candidate could be possibly inferred from its neighbors. The neighbors of a new drug candidate generally refer to those drugs that share some similar properties with the new drug candidate, e.g. similar in chemical structure.

For a drug candidate $d_i$ that has no known targets, we infer the weighted interaction profile for $d_i$
with the following formula

\begin{align}
\mathbf{l}(i)= \mathbf{s}^d_i \mathbf{A} \label{eq:infer}
\end{align}
where each dimension
\begin{align}
l_{j}(i)= \sum^{n_d}_{h=1} s^d_{ih}  a_{hj}
\end{align}
Here vector $\mathbf{s}^d_i$ is the $i$th column of $\mathbf{S}_d$, which records the similarities between $d_i$ and all the other drugs, $s^d_{ih}$ is the similarity between two drugs $d_i$ and $d_h$, and vector $\mathbf{l}(i)$ is the inferred interaction profile for $d_i$, where each dimension $l_{j}(i)$ corresponds to the weight of the interaction between $d_i$ and $t_j$.
The above formula shows that the interaction weight of $d_i$ with respect to the $j$th target is the sum of interactions between its neighbors and this target weighted by the similarity between this drug and its neighbors. More specifically, this simple formula defines that for a given new drug candidate $d_i$, its weight of interaction with respect to a target is high if many of its neighbors interact with this target, and the final weight to a target is influenced more by a neighbor with a larger similarity than those with smaller similarities. To allow neighbors with large similarities only to contribute, a threshold may be used to reduce the impact of those non-important neighbors to 0. Alternately, an exponential function with bandwidth $\beta$ given as below may be introduced:
\begin{align}
\mathbf{l}(i)= e^{(\mathbf{s}^d_i/\beta)}\mathbf{A}
\end{align}
To ensure the value of each $l_{j}(i)$ is in the range of [0, 1], linear scale is performed subsequently.
The procedure of inferring training data for new target candidates is not discussed in details here as it is similar to the procedure of inferring training data for new drug candidates as presented above.

Learning from neighbors allows drugs and targets to obtain training data when themselves do not have any known interactions. This procedure actually introduces some degree of globalization into the original local model to give more chances or an enlarged scope for the learning process. However, too much globalization is not desired as it will decrease the local characteristics and make the models for each drug or target less discriminative. Moreover, the low quality of neighbors may add in noise and cause a negative impact when neighbors' preferences are too much relied upon. In the current study, we only activate the neighbor-based training data inferring for totally new candidates. For other cases, we still train the model locally on its own preference, i.e., the known interactions.

\section{Experimental study}
Now we give some experimental results to compare the performance of the BGM method, the BLM method and the BLMN method for the task of drug-target interaction prediction. From the experimental results, we have the following observations: first, BLM-based approaches outperform BGM; second, with neighnor-based training data inferring, BLMN performs better than the classic BLM; third, network topology based similarity is helpful to improve the prediction.
\subsection{Datasets}
The four groups of datasets have been first analysed by \cite{Yamanishi08} and then later by several other researchers \cite{Bleakley09,Xia10,Laarhoven11,Chen12}. These four datasets correspond to drug-target interactions of four important categories of protein targets, namely enzyme, ion channel, G-protein-coupled receptor (GPCR) and nuclear receptor, respectively \footnote{The datasets were download from http://web.kuicr.kyoto-u.ac.jp/supp/yoshi/drugtarget/}. Table \ref{Tab:dataset} gives some statistics of each of the datasets.

Each dataset is described by three types of information in the form of three matrices. Together with the drug-target interaction information, the drug-drug similarity, and target-target similarity are also available. Four interaction networks were retrieved from the KEGG BRITE \cite{Kanehisa06}, BRENDA \cite{Schomburg04}, SuperTarget \cite{Gnther08} and DrugBank \cite{Wishart08} these four databases. The drug-drug similarity is measured based on chemical structures from the DRUG and COMPOUND sections in the KEGG LIGAND database \cite{Kanehisa06} and is calculated with SIMCOMP \cite{Hattori03}. The target-target similarity is measured based on the amio acid sequences from the KEGG GENEsS database \cite{Kanehisa06} and is calculated with a normalized version of Smith-Waterman score.
\begin{table}[!tbp]
\centering
\tbl{Some statistics of the four datasets. $n_d$: the total number of drugs, $n_t$: the total number of targets, $E$: the total number of interactions, $\bar{D}_d$: the average number of targets for each drug, $\bar{D}_t$: the average number of targeting drugs for each target, $D_d=1$: the percentage of drugs that have only one target, and $D_t=1$: the percentage of targets that have one targeting drug. }
{\begin{tabular}{lcccccccc}\hline
\footnotesize{Dataset} &\footnotesize{Enzyme}&\footnotesize{Ion Channel}&\footnotesize{GPCR}&\footnotesize{Nuclear Receptor}\\\hline
$n_d$  &445&210&  223              &54\\
 $n_t$ &664&204&  95             &26\\
 $E$ &2926&1476&635&90\\
 $\bar{D_d}$  &6.58   &7.03& 2.85  &1.67\\
 $\bar{D_t}$  &4.41   &7.24& 6.68  &3.46\\
 $D_d=1 (\%)$ &39.78&38.57&47.53&72.22\\
 %&177 &81& 106 &39\\
 $D_t=1 (\%)$ &43.37&11.27&35.79&30.77\\
 %&288  &23& 34&8\\

 \hline
 \end{tabular}}\label{Tab:dataset}

\end{table}

\subsection{Approaches compared}
We compare the following approaches:
\begin{itemize}
\item BGM \cite{Yamanishi08}: Bipartite graph model;
\item BY(2009) \cite{Bleakley09}: Bipartite local model;
\item Laarhoven et al (2011) \cite{Laarhoven11}: Bipartite local model with network-based; similarity
\item BLM: Ignoring `new candidate' in BLMN;
\item BLMN: BLM with neighbor-based training data inferring .
\end{itemize}
Among the above methods, BGM requires eigendecomposition of a $(n_c+n_d) \times (n_c+n_d)$ matrix, which is computational consuming for large datasets.  The BY(2009), Laarhoven et al (2011) and BLM are three variants of the classic BLM method, which is not applicable to new candidates. BLMN is the modified BLM method which can be used to predict the interaction between any compounds and proteins.
\subsection{Evaluation}
  Leave-one-out cross validation (LOOCV) is performed. In each run of prediction, one drug-target pair is left out by setting the corresponding entry of matrix $\mathbf{A}$ to 0. Then we try to recover its true value using the remaining data.
We measure the quality of the predicted interaction matrix $\mathbf{P}$ by comparing it to the true interaction matrix $\mathbf{A}$ in terms of the area under ROC curve or true positive rate (TPR) vs. false positive rate (FPR) curve (AUC) and the area under the precision vs. recall curve (AUPR). TPR is equivalent to recall. Assume that TP, FP, TN, FN represent true positive, false positive, true negative, and false negative, respectively, then
\begin{align}
TPR/recall=\frac{TP}{TP+FN}\\
FPR=\frac{FP}{FP+TN}\\
precision=\frac{TP}{TP+FP}
\end{align}
Since in the current task, the known interactions are much less than those unknown ones, the precision-recall curve should be a better measurement than the ROC curve here as has been discussed in \cite{Davis06}.

\subsection{Performance comparison}
 Table \ref{Tab:results_comp} gives the AUC and AUPR scores of five approaches on the four datasets. The results of BGM, BY (2009), and Laarhoven et al (2011) are the best ones reported in \cite{Bleakley09} and \cite{Laarhoven11}. Both BLMN and BLM are run with three different groups of inputs: \emph{Chem-Seq}, \emph{Network-based}, and \emph{Hybrid}. \emph{Chem-Seq} denotes that chemical similarity is used for drug and sequence similarity is used for target; \emph{Network-based} denotes that the drug-drug similarity and target-target similarity are derived from the existing interaction network; \emph{Hybrid} denotes that the drug-drug similarity and target-target similarity are combinations of the two types of similarities.

It is shown from the table that with a low time complexity, four BLM-based approaches, including three BLM variants and BLMN, produce better results than the BGM method. Among the three BLM variants, the results of BLM and BY(2009) with \emph{Chem-Seq} are similar as the only difference between them is the former use RSL as the classifier while the later use SVM. The results of BLM and Laarhoven et al (2011) with \emph{Network-based} are also close in most of the cases although the later used Kronecker product, which is a more complicated way to combine two types of similarities. In all the cases, BLMN produced better results than the three classic BLM algorithms. This clearly show that neighbor-based training data inferring is very useful for improving the final result when the dataset contains new drug/target candidates.

Despite the consistent improvements of BLMN compared to the other three on all the four datasets, the amounts of improvements differ for different datasets. If we compare the improvements of the proposed approaches over the four datasets, it is seen that the improvement with respect to BLM on Nuclear Receptor is the most significant while the improvement on Enzyme and Ion Channel are not so significant. Such kind of differences in performance of the proposed approach are consistent with our expectation according to the differences in the structure of the datasets. Although all the datasets do not contain new drug/target candidates, in our experiment, the real interaction to be predicted is leave out. This means drugs and targets with degree equal to 1 turn out to have no positive training data and thus they are simulated to be ``new'' in the experiments. As shown in Table \ref{Tab:dataset}, Nuclear Receptor has a much larger portion of ``new'' drugs and targets than Ion Channel. Therefore, it has more chances for BLMN to improve the results for Nuclear Receptor where the training data inferring is applied more frequently.

It is also observed that although network-derived similarity alone provides good information, combining biological information can further improves the result especially when the network is sparse, e.g., the results of both BLM and BLMN for Ion Channel with only \emph{Network-based}is very close to those with \emph{Hybrid} while significant improvements are achieved for both approaches on Nuclear Receptor when \emph{Chem-Seq} is further combined with \emph{Network-based} similarity. This shows that combining multiple types of similarities usually gives better results when no single type of similarity is good enough.

\begin{table}[!t]
\tbl{Comparison of AUC and AUPR for the four datasets}
{\begin{tabular}{lllll}\hline Dataset &Data&Method &AUC
&AUPR\\\hline
 Enzyme  & \emph{Chem-Seq}
 & BGM&96.7&83.1\\
 && BY(2009) &97.6&83.3\\
 && BLM&96.1 &85.8\\
 &       &BLMN &98.0&87.3\\  \cline{2-5}
         &  \emph{Network-based}& Laarhoven et al (2011) & 98.3 &88.5\\
         & &BLM     &98.2  &88.0\\
        &       &BLMN  &99.1  &93.1\\  \cline{2-5}
        & \emph{Hybrid}   & Laarhoven et al (2011) &97.8&91.5\\
        &&BLM   &98.2 & 91.3 \\
        &       &BLMN &98.8  & 92.9\\  \hline
Ion Channel& \emph{Chem-Seq}
& BGM&96.9&77.8\\
&& BY(2009) &97.3&78.1\\
&&BLM &97.0&81.9\\
 &       &BLMN &97.8&84.6\\  \cline{2-5}
           &\emph{Network-based}&Laarhoven et al (2011)&98.6&92.7\\
           &&BLM   &98.5  &92.5\\
            &       &BLMN  &99.0  &95.6\\\cline{2-5}
            &\emph{Hybrid} &Laarhoven et al (2011)&98.4&94.3\\
            &&BLM            &98.5  &92.7\\
        &       &BLMN     &99.0  &95.0\\ \hline
 GPCR    & \emph{Chem-Seq}
 & BGM&94.7&66.4\\
 && BY(2009) &95.5&66.7\\
 &&BLM &95.1&68.1\\
 &       &BLMN &98.1&78.8\\  \cline{2-5}
         & \emph{Network-based}&Laarhoven et al (2011)&94.7&71.3\\
         &&BLM    &94.4  &70.6\\
         &     &BLMN  &97.5  &84.6\\\cline{2-5}
       &\emph{Hybrid}&Laarhoven et al (2011)&95.4&79.0\\
       &&BLM         &95.7  &76.2\\
       &       &BLMN   &98.4 &86.5\\ \hline
Nuclear Receptor & \emph{Chem-Seq}
& BGM&86.7&61.0\\
&& BY(2009) &88.1&61.2\\
&&BLM &86.9& 58.4\\
 &       &BLMN &96.9&80.7\\  \cline{2-5}
           & \emph{Network-based}&Laarhoven et al (2011)&90.6&61.0\\
           && BLM&90.9  &62.9\\
             &&BLMN          &95.7  &80.7\\\cline{2-5}
         &\emph{Hybrid}  &Laarhoven et al (2011)&92.2&68.4\\
         &&BLM   &94.0  &72.4\\
    &&BLMN   &98.1   &86.6\\
\hline
 \end{tabular}}\label{Tab:results_comp}
\end{table}

\section{Summary}
Classification is an important data analysis tool that have been studied extensively. Many computational biology tasks are binary classification problem that predicts the outcome of a trial is positive or negative. We have introduced several popular supervised learning methods for classification including popular classification methods, regression models used for classification, and ensemble classification. We give more detailed discussion of how  different classification methods can be used for drug-target interaction prediction. Experimental studies are given to compare the performance of different approaches with benchmark datasets.

Other than the specific learning method, the classification result is also highly dependent on the amount and quality of the given training data and the way the data represented, e.g., a set of features or similarity measures. Given the same set of training data, a good data representation with a simple classifier may already produces a good result. Nevertheless, with the same data representation, an advanced classification algorithm is able to make use of it more effectively and hence produce a better result. This chapter focus on algorithm design.

\bibliographystyle{ws-rv-van}
\bibliography{classification_app}

%\printindex[aindx]                 % to print author index
\printindex                         % to print subject index
\end{document}